\pdfoutput=1
\documentclass[11pt]{article}

\usepackage[final]{acl}
\usepackage{xspace}
\usepackage{booktabs}
\usepackage{comment}
\usepackage{pifont}  % For \checkmark (\cmark) and \ding{55} (\xmark)
\usepackage{multirow}

\newcommand{\modelname}{DP-GTR\xspace}
\usepackage{amsmath}
\usepackage{amssymb}
\usepackage{algorithm}
\usepackage{algorithmic}
\usepackage{pifont}
\usepackage{booktabs}

\usepackage{subcaption}

\usepackage{makecell}
\usepackage{setspace}
\usepackage{enumitem}
\definecolor{green}{rgb}{0,0.5,0}
\definecolor{red}{rgb}{0.8,0,0}
\definecolor{updateColor}{rgb}{0,0,0.5}
\definecolor{mred}{RGB}{210, 70, 80} 
\definecolor{mgreen}{RGB}{60, 160, 100}
\definecolor{mybrown}{RGB}{106,94,57}
\definecolor{myyello}{RGB}{253,246,221}

\newcommand{\mypara}[1]{\noindent{\bf {#1}.}\xspace}
\usepackage{tcolorbox} 
\usepackage{cuted}
\usepackage{cleveref}
\newcounter{mybox}

\usepackage{times}
\usepackage{latexsym}

\usepackage[T1]{fontenc}
\usepackage[utf8]{inputenc}

\usepackage{microtype}

\usepackage{inconsolata}

\usepackage{graphicx}

\title{\modelname: Differentially Private Prompt Protection via Group Text Rewriting}

\author{
Mingchen Li\textsuperscript{$1,\spadesuit$},
Heng Fan\textsuperscript{$1,\diamondsuit$},
Song Fu\textsuperscript{$1,\diamondsuit$},
Junhua Ding\textsuperscript{$2,\diamondsuit$},
Yunhe Feng\textsuperscript{$1,\diamondsuit$}\\
University of North Texas\\
$^{1}$ Department of Computer Science and Engineering \quad
$^{2}$ Department of Data Science\\
$\spadesuit$ \texttt{MingchenLi@my.unt.edu} \\
$\diamondsuit$ \texttt{\{heng.fan, song.fu, junhua.ding, yunhe.feng\}@unt.edu}
}

\begin{document}
\maketitle
\begin{abstract}
Prompt privacy is crucial, especially when using online large language models (LLMs), due to the sensitive information often contained within prompts. While LLMs can enhance prompt privacy through text rewriting, existing methods primarily focus on document-level rewriting, neglecting the rich, multi-granular representations of text. This limitation restricts LLM utilization to specific tasks, overlooking their generalization and in-context learning capabilities, thus hindering practical application. To address this gap, we introduce \modelname, a novel three-stage framework that leverages local differential privacy (DP) and the composition theorem via \textit{group text rewriting}. \modelname is the first framework to integrate both document-level and word-level information while exploiting in-context learning to simultaneously improve privacy and utility, effectively bridging local and global DP mechanisms at the individual data point level. Experiments on CommonSense QA and DocVQA demonstrate that \modelname outperforms existing approaches, achieving a superior privacy-utility trade-off. Furthermore, our framework is compatible with existing rewriting techniques, serving as a plug-in to enhance privacy protection. Our code is publicly available at \href{https://github.com/ResponsibleAILab/DP-GTR}{github.com/ResponsibleAILab/DP-GTR}.
\end{abstract}

% 卖点: (优先级排序)
% 1. 当前方法对于转述文档的使用不足, 通过GTR, 通过 ICL, 同时控制 utility 和 privacy, 不再单一使用 \epsilon; 对于隐私, 是 priavte keywords 生成抑制; 对于 utility, 是 ICL 1-shot.
% 2. 通过 GTR, 在 local 中整合出来 global 的操作, 桥接了localDP与 globalDP的机制可用.
% 3. 模拟了更加 practical QA scenario, 实验比 author obfuscation 更加实际, 细粒度.
% 4. 作为插件兼容所有的 paraphrasing 办法, 并能够有效提高

\section{Introduction}
% The rise of LLMs in natural language processing has spurred research into differential privacy (DP) techniques to mitigate the risk of sensitive information leakage \citep{dpsgd, dpicl, dpicl2}. 
The rise of LLMs in natural language processing has catalyzed an urgent research focus on their security and privacy~\cite{yao2024survey, peng2025jalmbench, luo2025unsafe}, including investigations into differential privacy (DP) techniques to mitigate the risk of sensitive information leakage~\citep{edemacu2025privacy, dpsgd, dpicl, dpicl2}.
While DP, the gold standard for computational privacy, has seen broad adoption in machine learning, existing text-based DP methods face significant challenges. These methods generally fall into four categories: training-based optimizations (e.g., DP-SGD~\citep{dpsgd}), embedding perturbations~\citep{feyisetan2020privacy}, document-level paraphrasing~\citep{dpparaphrase}, and in-context learning (ICL) enhancements~\citep{dpicl}.  However, training-based approaches are computationally expensive, embedding perturbations can compromise semantic coherence, and ICL often neglects client-side prompt privacy.

Document paraphrasing offers a promising balance between privacy and utility. State-of-the-art methods achieve differentially private next-token generation using the exponential mechanism (EM)~\citep{mcsherry2007em, carvalho2023tem}, replacing the standard softmax.  Initial work employed decoder-only models like fine-tuned GPT-2~\citep{dpparaphrase}, progressing to encoder-decoder~\citep{igamberdiev2023dpbart} and encoder-only architectures (e.g., BART, RoBERTa)~\citep{dpmlm}.  DP-Prompt~\citep{dpprompt} leverages prompt learning for zero-shot paraphrasing, and recent advancements combine DP post-processing with adversarial fine-tuning~\citep{meisenbacher2024postprocessing}.  A critical limitation, however, persists: lack of fine-grained control over the privacy-utility trade-off.

Current EM-based DP methods provide only coarse-grained control via the privacy budget, hindering practical deployment.  Most approaches (except DP-Prompt) also necessitate resource-intensive fine-tuning. While document-level paraphrasing preserves more contextual information than embedding perturbations, it often overlooks word-level privacy vulnerabilities. These limitations highlight the need for a training-free, fine-grained privacy solution that fully leverages textual information, a capability well-suited to the ICL paradigm of LLMs.

Prior work on DP in ICL has predominantly focused on server-side, global DP implementations, often using a "sample-and-aggregate" approach~\citep{nissim2007sampleandagg} to privately partition and aggregate context databases~\citep{dpicl,dpicl2}.  Client-side prompt privatization, in contrast, requires the stronger guarantees of local DP (LDP), protecting individual data points rather than entire datasets.  This distinction creates a significant gap between global and local DP in ICL, motivating the need for approaches that bridge it.

\begin{figure*}[t]
    \centering
    \includegraphics[width=\linewidth]{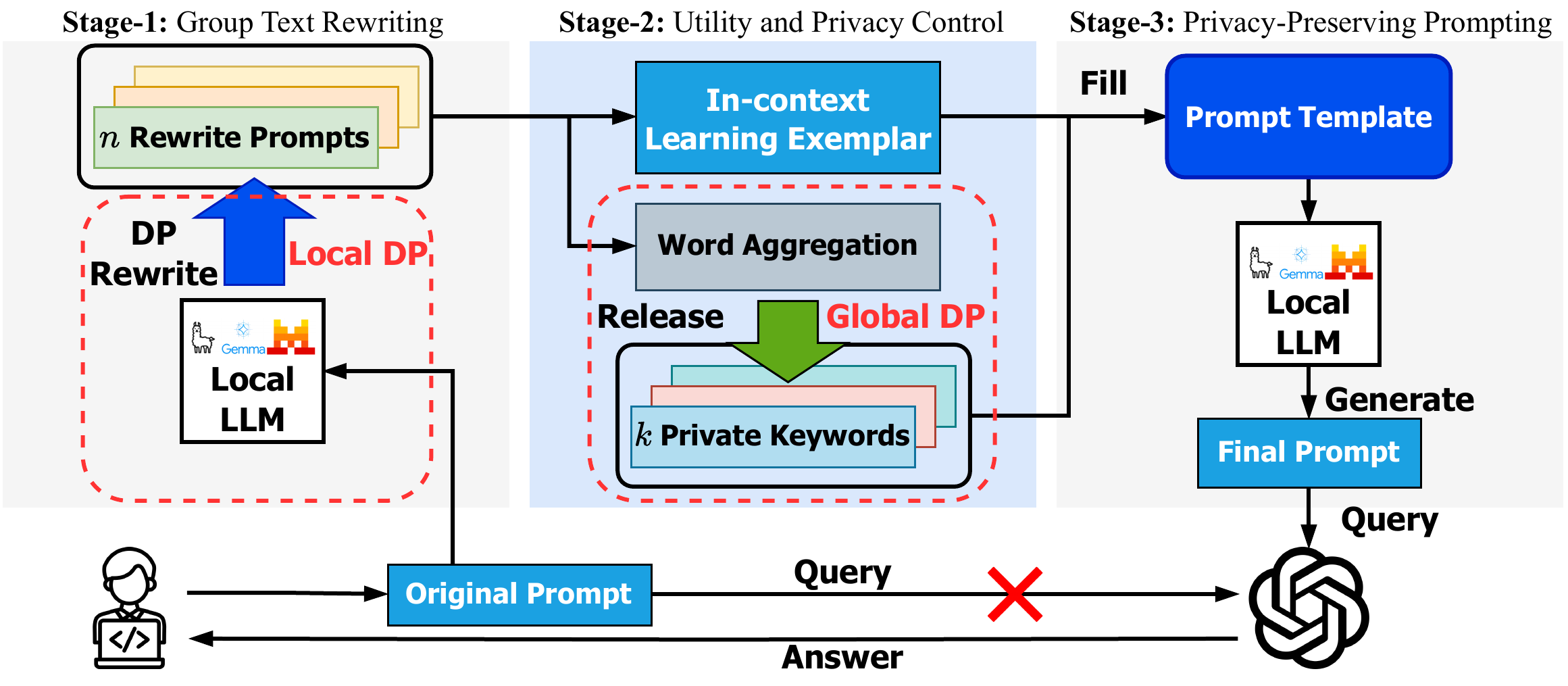}
    \caption{\modelname: A three-stage pipeline for \underline{D}ifferentially \underline{P}rivate prompt protection via \underline{G}roup \underline{T}ext \underline{R}ewriting (GTR). Stage-1 generates \(n\) paraphrases of the original prompt using a DP paraphrasing mechanism. Stage-2 identifies the lowest-perplexity prompt as the ICL exemplar and aggregates word counts to release \(k\) private keywords. Stage-3 integrates these private keywords and the ICL exemplar into a prompt template for submission to the LLM, producing the final, differentially private prompt.}
    \label{fig:framework}
\end{figure*}

Addressing these gaps, we propose \modelname, a three-stage, differentially private prompt protection framework built upon a novel \textit{Group Text Rewriting (GTR)} mechanism (see Figure~\ref{fig:framework}). \modelname is designed to provide fine-grained control over the privacy-utility trade-off while remaining compatible with existing paraphrasing techniques. In Stage-1, GTR generates multiple client-side paraphrases of an input prompt, forming a "rewriting group" that preserves rich contextual information and enables bag-of-words-like count analysis. Notably, GTR connects local and global DP principles on the client-side. Stage-2 uses these counts for fine-grained privacy-utility control, identifying potentially sensitive private consensus keywords – words appearing frequently across paraphrases despite DP-driven variations. We mitigate this risk by releasing a fixed number of these keywords or using a differentially private aggregator, and select the lowest-perplexity paraphrase to maximize output quality. Stage-3 suppresses the identified keywords, limiting privacy leakage, and uses the selected paraphrase as an ICL example to improve utility. In addition, we evaluate \modelname in a realistic question-answering (QA) scenario, simulating real-world LLM usage.

Our key contributions are:

\begin{itemize}[nosep]
    \item We propose \textit{Group Text Rewriting (GTR)}, a novel mechanism bridging local and global DP at the client-side prompt, enabling the integration of various DP techniques.
    \item We present \modelname, a three-stage prompt protection framework leveraging ICL for fine-grained privacy-utility control, compatible with existing paraphrasing methods.
    \item To our knowledge, we are the first to unify document-level and word-level privacy considerations within a single framework.
    \item We evaluate state-of-the-art DP paraphrasing methods in a realistic QA setting, demonstrating \modelname's superior privacy-utility trade-off compared to existing approaches.
\end{itemize}

\section{Related Work}  % 0.75页, 到第二页结束.
\textbf{Global vs. Local DP:} Differential privacy text sanitization methods are classified into Global Differential Privacy (Global-DP) and Local Differential Privacy (Local-DP) based on where the privacy mechanism is applied. In Global-DP, data is aggregated centrally before applying the privacy mechanism, while methods like DP-SGD use differentially private optimization techniques for training text models~\citep{dpsgd, ponomareva2022training, feyisetan2020privacy}. DP-ICL operates within a ``sample-and-aggregate'' framework by perturbing the embedding and vocabulary selected for release~\citep{dpicl}. In contrast, Local-DP incorporates the differential privacy mechanism before data reaches the centralized processor, typically affording stronger privacy protection~\citep{duchi2013local, feyisetan2020privacy}.

\noindent\textbf{Local-DP:} Private document release methods are categorized into three tiers based on where noise is added: word-level, sentence-level, and document-level. At the word level, noise is added to word embeddings, and the perturbed vectors are then mapped to the nearest vocabulary word~\citep{feyisetan2020privacy,xu2020differentially,yue2021differential}. ~\citet{carvalho2023tem} employ the exponential mechanism for token selection, while ~\citet{cuztext} propose customized token mappings for individual words. Moreover, ~\citeposs{meisenbacher20241} study generates multiple candidate perturbations using various word embedding models, and ~\citet{dpparaphrase} highlights that word-level approaches inherently lack contextual information. Similarly, sentence-level methods inject noise into sentence embeddings ~\citep{reimers2019sentencetrans,meehan2022sentencedp}.

\noindent\textbf{Document-level:} At the document level, paraphrasing technologies are grouped into three categories based on model architecture. \citet{dpparaphrase} uses a decoder-only fine-tuned GPT-2 model. Later work adopts encoder-decoder models, such as a BART-based approach~\citep{lewis2019bart} with sensitivity clipping via thresholding and pruning~\citep{igamberdiev2023dpbart}. Encoder-only methods, like DP-MLM~\citep{dpmlm}, use a RoBERTa-based masked language model for fine-tuning. These methods require fine-tuning. In contrast, DP-Prompt~\citep{dpprompt} introduces a zero-shot prompt learning paradigm using black-box LLMs, and \citet{meisenbacher2024postprocessing} employs post-processing and adversarial fine-tuning to enhance rewriting.

\noindent\textbf{DP in ICL:} The primary concern with applying DP in ICL is that LLMs are not inherently secure, potentially exposing sensitive context. \citet{dpicl}'s DP-ICL perturbs embeddings and extracts keywords from subsampled datasets, while \citet{dpicl2} incorporates label information. \citet{zheng2024locallyicl} employ \(k\)-RR~\citep{wang2017locallyprotocols} to generate ICL answers, and \citet{gao2024Data-adaptiveicl} aggregate next-token predictions from dataset shards. All the approaches follow a ``sample-and-aggregate'' framework~\citep{nissim2007sampleandagg}, partitioning the data and applying private aggregation.

Our work, \modelname, draws on the principles of prompt learning and the ``sample-and-aggregate'' strategy from DP-Prompt and DP in ICL respectively. This one-shot in-context learning framework, analogous to global DP, obviates resource-intensive fine-tuning while enhancing both privacy protection and utility.

\section{Preliminaries} 
\label{sec:prelim}
\mypara{Threat Model} We assume an adversary attacker on the side of the cloud-based model vendor, whose objective is to extract private information (such as personal identifiable information) from the confidential transmission content. The adversary’s access is limited to a customized prompt supplied by the client, but they are free to facilitate their attack.

\paragraph{Pure Differential Privacy (DP)} A randomized mechanism $\mathcal{M}: \mathcal{X} \to V$ satisfies $\epsilon$-Pure DP if, for any neighboring datasets $D$ and $D'$ differing by at most one element, and any output $V \subseteq \text{Range}(\mathcal{M})$, the following inequality holds $\Pr[\mathcal{M}(D) = V] \leq e^{\epsilon} \cdot \Pr[\mathcal{M}(D') = V]$ ~\citep{dwork2006calibrating}.

\paragraph{Local Differential Privacy} Local DP applies a mechanism $\mathcal{M}$ to each individual data point $x, x' \in \mathcal{X}$ (where $x$ and $x'$ are considered neighboring in some sense), generating a local perturbation $V$ before the data is submitted to the data center ~\citep{duchi2013local, dwork2006calibrating}.

\paragraph{Metric Differential Privacy} To improve the utility of DP, the indistinguishability of two outputs for $x$ and $x'$ can be scaled by the distance between their corresponding inputs ~\citep{alvim2018metric}. A mechanism $\mathcal{M}$ satisfies $\epsilon$-Metric DP if, for any inputs $x, x' \in \mathcal{X}$ and any output $V \subseteq \text{Range}(\mathcal{M})$, the following inequality holds:
\[
\Pr[\mathcal{M}(x) = V] \leq e^{\epsilon \cdot d(x,x')} \cdot \Pr[\mathcal{M}(x') = V],
\]
where $d(x, x')$ is a distance metric defined on $\mathcal{X}$.

\paragraph{Exponential Mechanism} The \emph{Exponential Mechanism} (EM) injects noise into scoring functions, making it suitable for non-numeric sensitive queries ~\citep{mcsherry2007em}. Given a dataset $D$ and a utility function $u: D \to V$, where $V$ is the set of possible outputs, the mechanism $\mathcal{M}$ is defined as
\[
\Pr[\mathcal{M}(D) = v] \propto \exp\left(\frac{\epsilon\, u(D,v)}{2\, \Delta u}\right),
\]
where the sensitivity $\Delta u$ is defined as
\[
\Delta u = \max_{D,D',v} \bigl| u(D,v) - u(D',v) \bigr|,
\]
and the maximum is taken over all neighboring datasets $D$ and $D'$ and all possible outputs $v \in V$.

\paragraph{Composition Property} Differential privacy exhibits a robust \emph{composition property}: when multiple DP mechanisms are applied sequentially to the same dataset, the overall privacy loss accumulates~\citep{dwork2014algorithmic}. Let $D$ be a dataset and let $M_1, M_2, \dots, M_n$ be $\epsilon_i$-DP mechanisms. The composed mechanism
\[
M = M_n \circ M_{n-1} \circ \cdots \circ M_1
\]
satisfies $\epsilon$-DP with $\epsilon = \sum_{i=1}^{n} \epsilon_i$.

\paragraph{Post-Processing Property} The \emph{post-processing property} states that any function applied to the output of a DP mechanism preserves the same privacy guarantee~\citep{dwork2014algorithmic}. If a mechanism $\mathcal{M}: \mathcal{X} \to V$ satisfies $\epsilon$-DP, then for any function $F: V \to V'$, the composed mechanism $F \circ \mathcal{M}(D)$ also satisfies $\epsilon$-DP.

\paragraph{DP-Guaranteed Paraphrasing} Autoregressive language models (LMs) generate text \textit{sequentially}, sampling tokens from a conditional likelihood distribution: $\prod_{i=1}^{n} \Pr[x_{i} \,|\, x_{1},\dots, x_{i-1},\, C]$, where $C = (c_{1}, c_{2}, \dots, c_{m})$ is the context.  At each step, a logit vector $u \in \mathbb{R}^{|\mathcal{V}|}$ is transformed into a probability distribution over the vocabulary $\mathcal{V}$ using a softmax function with temperature $T$:
\[
p(v) = \frac{\exp(u_{v}/T)}{\sum_{w \in \mathcal{V}} \exp(u_{w}/T)}, \quad \forall\,v \in \mathcal{V}.
\]
Prior work ~\citep{dpprompt, dpparaphrase} has shown the equivalence between this softmax selection process and the Exponential Mechanism (EM) of differential privacy, where the utility function corresponds to the logits. Assuming $LM$ is not pre-trained on the distribution of the data being protected, and that logits $u_v$ are clipped to $[b_{\min}, b_{\max}]$, generating $n$ tokens at temperature $T$ provides a $(\frac{2n(b_{\max}-b_{\min})}{T})$-local DP (LDP) guarantee. This derives from the fact that each token selection, with a maximum logit difference of $(b_{\max} - b_{\min})$, incurs a privacy loss of $\frac{2(b_{\max} - b_{\min})}{T}$. Sequential composition over $n$ tokens then yields the stated LDP bound for a single document paraphrase. Logit clipping and EM sampling ensure the generated sequence respects a well-defined pure LDP budget. See Appendix~\ref{sec:proof} for details.

\section{\modelname}
Existing document-level prompt sanitization methods often employ the EM for privacy-preserving rewriting. However, these coarse-grained approaches, relying on a single $\epsilon$ for the entire document (prompt) rewriting, struggle to balance privacy and utility. Critically, noise introduced during rewriting irreversibly alters textual elements, hindering utility recovery. Maintaining acceptable utility thus necessitates low initial noise levels, requiring a high privacy budget and consequently reducing actual privacy protection. Furthermore, a high privacy budget under the EM can even lead to complete data exposure. To address these limitations, we propose \modelname, a word- and document-level hybrid prompt privacy adopted framework that leverages group text rewriting and post-processing to enhance privacy while maintaining high utility under DP guarantees. \modelname enables low-noise paraphrasing to identify and suppress the generation of privacy-sensitive terms with high exposure.

\subsection{\modelname Framework Overview}
\modelname comprises three distinct stages, as illustrated in Figure~\ref{fig:framework}. In Stage-1, a DP-guaranteed group text rewriting process explores diverse representations of the original prompt, generating $n$ rewritten versions. Stage-2 leverages this group of rewritten prompts in a parallel process. First, it identifies the lowest-perplexity rewritten prompt as an in-context learning exemplar, effectively selecting the most confident paraphrase. Concurrently, it aggregates word counts across the rewritten prompts and releases $k$ private keywords shared within the group. Finally, Stage-3 employs a prompt template. This template is populated with both the selected in-context learning exemplar and the released private keywords. The filled template is then fed to the LLM to generate the final prompt, effectively mitigating the risk of directly revealing sensitive information from the original prompt.

\subsection{Stage-1: Group Text Rewriting}
\modelname employs group text rewriting to achieve finer-grained control over privacy and utility compared to document-level methods.  Effective prompt sanitization requires considering both document-level context for overall meaning and word-level information for protecting sensitive terms. While LLMs excel with contextual input, directly using the original prompt compromises privacy.

Group text rewriting addresses this by generating a local paraphrased text database, effectively mitigating the limitations of both document-level rewriting and the absence of suitable contextual information. This database, consisting of multiple rewritten versions of the prompt, serves several key purposes. First, it provides richer information than a single rewrite, capturing diverse facets of the original prompt. Second, aggregating word counts across the rewrites facilitates the identification of shared, potentially sensitive keywords. More importantly, the group enables more aggressive post-processing and additional DP mechanisms, which, while not reducing the formal $\epsilon$-DP bound, further mitigate the risk of sensitive information disclosure by eliminating sensitive identifiers and limiting real-world exposure.
Specifically, generating a paraphrased group $\mathcal{P}$ of $m$ documents, each with $n$ tokens, incurs an $(mn)\epsilon_1$-DP privacy budget.

\subsection{Stage-2: Utility and Privacy Control}
\modelname achieves fine-grained control over utility and privacy by leveraging the group of rewritten prompts.  Utility is enhanced through in-context learning, while privacy is preserved via private keyword analysis.

\subsubsection{One-shot in-context learning for utility}

LLMs exhibit a strong capacity for in-context learning ~\citep{icl}, effectively learning from provided examples. To ensure the LLM understands the desired output format and content, contextual information is crucial. Furthermore, LLMs often demonstrate a preference for learning from their own generated content. Therefore, we employ a one-shot in-context learning approach to maximize the utility of the rewritten prompt.

Specifically, rather than using the original prompt, we select the lowest-perplexity paraphrase, $P_{\text{low}}$, from the generated group $\mathcal{P}$ as the exemplar for guiding final prompt generation.  This paraphrase, representing the most coherent and representative information within the group, serves as the most effective learning example. Critically, this in-context learning process leverages the post-processing property of differential privacy, incurring no additional privacy budget. Formally, this can be seen as the temperature approaching zero as the privacy budget approaches infinity: $T = \lim_{\epsilon \to \infty} \frac{2(b_{\max}-b_{\min})}{\epsilon} \rightarrow 0$.

\subsubsection{Consensus-aware privacy protection}

Protecting prompt privacy requires identifying privacy-sensitive keywords. Unlike previous PII detection methods that focus on isolated words or phrases ~\citep{has, cuztext}, \modelname considers the overall \emph{composition} of sentences to comprehensively capture privacy leakage risks. Due to the paraphrasing tendencies of LLMs, key pieces of information within these compositional relationships often reappear across different paraphrased examples. We define \textit{consensus words} as those that appear repeatedly across multiple paraphrased prompts generated in Stage-1. This repetition is treated as a privacy signal, as words appearing frequently despite paraphrasing attempts are likely either (a) crucial to the document's meaning and difficult to alter without significant utility loss, or (b) inherently tied to sensitive or identifiable information (e.g., names, locations) that existing LDP methods struggle to effectively anonymize. 

\noindent
\textbf{Consensus Keyword Extraction}
The paraphrased group generated in Stage-1 can reproduce large fragments, potentially "leaking" sensitive information. Inspired by the bag-of-words approach, we count word frequency ($c$) across the paraphrased sentences, forming a set of frequency counts $\mathcal{S} = \{(w_1, c_1), (w_2, c_2), \dots, (w_k, c_k)\}$. We then release a fixed number ($K$) of keywords, $\mathcal{K} = \{k_1, k_2, \dots, k_K\}$, without manual intervention. This can be achieved either through post-processing or by employing the Joint Exponential Mechanism (Joint-EM) ~\citep{gillenwater2022jointEM} with privacy budget $\epsilon_2$ under the sample-and-aggregate framework ~\citep{nissim2007smooth}. The consensus keyword extraction algorithm is detailed in Algorithm~\ref{alg:topk_extraction}.

\noindent
\textbf{Post-Processing Release} Due to the post-processing property of DP, keywords $\mathcal{K}$ can be directly released, incurring no additional differential privacy (NDP) budget. This allows for diverse downstream analyses without further privacy cost, maintaining the total budget at $(mn)\epsilon_1$-LDP. This approach is designated \textbf{\textit{\modelname-NDP}}.

\noindent
\textbf{Joint-EM Release} Joint-EM provides a privacy-preserving DP mechanism for simultaneously releasing the top-$K$ keywords ~\citep{gillenwater2022jointEM}, making it a suitable alternative.  This approach has a total privacy budget of $((mn)\epsilon_1 + \epsilon_2)$-LDP and is designated \textbf{\textit{\modelname-JEM}}.

\begin{algorithm}[ht]
% \setstretch{0.85}
\caption{Top-$K$ Private Keywords Extraction}
\label{alg:topk_extraction}
\begin{algorithmic}[1]

\REQUIRE $\{P_1, P_2, \ldots, P_M\}$: paraphrased documents; $K$: output word count; \texttt{method} $\in \{\texttt{post-processing}, \texttt{Joint-EM}\}$; privacy budget $\epsilon$

\ENSURE Top-$K$ highest-frequency words

\FOR{$i \gets 1$ to $M$}
  \STATE $S_i \gets \textsc{SeparateBySpace}(P_i)$
  \STATE $S_i \gets \textsc{RemoveStopWords}(S_i)$
  \STATE \texttt{private\_keywords} $\gets \{\}$

  \FORALL{$w \in S_i$}
    \STATE \texttt{private\_keywords}$[w] \texttt{ += } 1$
  \ENDFOR

  \STATE \textsc{SortDescending}(\texttt{private\_keywords})
\ENDFOR

\IF{\texttt{method} = \texttt{post-processing}}
  \RETURN \textsc{TopK}(\texttt{private\_keywords}, $K$)
\ELSIF{\texttt{method} = \texttt{Joint-EM}}
  \RETURN \textsc{JointEM-TopK}\\  
\hspace{3em} (\texttt{private\_keywords},$\epsilon$, $K$)
\ENDIF

\end{algorithmic}
\end{algorithm}

\subsection{Stage-3: Privacy-Preserving Prompting}

To maximize utility while preserving prompt privacy, \modelname constructs the final prompt in Stage-3. Leveraging LLM prompt learning, particularly the stronger learning aptitude exhibited with negative commands~\citep{zhong2024rose, wei2022chain}, we utilize the extracted consensus keywords to effectively \emph{prevent} the generation of private information. This approach offers both practical and gentle privacy protection. Practically, we directly instruct the LLM to avoid generating the identified private keywords, eliminating the need for further word, token, or document modification, thus streamlining the process. Gentle privacy is achieved by strategically engineering prompts to selectively suppress model output, rather than relying on simple filtering rules or context-agnostic direct replacement. 

To maximize utility, we incorporate the lowest-perplexity rewritten prompt, $P_{\text{low}}$, selected in Stage-2, as a one-shot in-context learning example. Simultaneously, to ensure privacy, we instruct the LLM to avoid generating the private keywords, $w_1, w_2, \dots, w_k$, released in Stage-2. Our extracted keywords contain richer combinatorial information and global context, enabling this more nuanced control compared to other methods. The resulting prompt template is shown below.

\begin{center}
    \begin{tcolorbox}[
        colback=gray!5, 
        colframe=black, 
        boxrule=1pt, 
        rounded corners,
        title=\textbf{Privacy-Preserving Prompt Template}, 
        fonttitle=\bfseries,
        boxsep=1pt,            % Reduces padding inside the box
        left=2pt, right=2pt, top=2pt, bottom=2pt  % Tightens margins inside the box
    ]
    Refer to the following question to generate a new question:  
    <\(P_{low}\)>  
    Avoid using the following tokens:  
    <\(w_1, w_2, ..., w_k\)>
    \end{tcolorbox}
\end{center}

\section{Experiment}
\subsection{Limitations of Current Metrics}
Prior work on prompt privacy preservation, primarily focused on author obfuscation \citep{dpprompt} using datasets like \textit{Yelp} and \textit{IMDb}, typically evaluates privacy via adversarial classifiers attempting to identify the original author and utility through binary sentiment classification using BERT-based models \citep{kenton2019bert}. These evaluations, however, are often coarse-grained and fail to capture nuanced changes in meaning or style. For instance, a severely degraded paraphrase like "!!!!!" might be deemed to protect the author's identity and preserve the original positive sentiment of "At least for me, this movie is good!!," despite a significant loss of information, bordering on hallucination. Such "protection," arising from factors like high temperature settings, specific formatting, autoregressive generation, or model limitations, highlights the inadequacy of these existing evaluation metrics for assessing real-world applicability, as they fail to penalize extreme modifications that compromise the prompt's informational content.

\subsection{Experiment Setups}
\label{sec:exp_setup}
To evaluate prompt privacy and utility in a practical LLM service context, we propose an integrated question answering (QA) evaluation framework, conducting a single QA round to simultaneously measure both privacy and security. We use two QA datasets: the 5-choice closed-answer Commonsense QA (\textit{\textbf{CSQA}}) \citep{csqa} and the open-answer PFL-DocVQA (\textit{\textbf{VQA}}) \citep{docvqa}, selecting 200 random items from each dataset's validation set. Note that VQA provides pre-extracted OCR tokens.

\noindent\textbf{Integrated Evaluation.} We simultaneously evaluate prompt privacy and utility. Privacy is measured by minimizing Rouge1, RougeL \citep{lin2004rouge}, and BLEU \citep{papineni2002bleu} scores between the original prompt \(p\) and the sanitized prompt \(p'\), indicating \textit{\textbf{greater privacy with more dissimilar prompts}}. Utility is assessed using a GPT-3.5 \citep{GPT35} based evaluator: Accuracy for the closed-answer dataset (CSQA) and Rouge1 for the open-answer dataset (VQA), comparing the LLM's answer \(a\) (generated from \(p'\)) to the ground truth. \textit{\textbf{Lower similarity scores indicate better privacy, while higher accuracy/Rouge1 scores indicate better utility}}.

\noindent\textbf{Comparative Baselines.} We employ three competitive approaches as baselines: DP-Prompt \citep{dpprompt}, a strong baseline leveraging zero-shot prompt learning on LLMs (GPT-3.5, Llama-3.1-8B \citep{Llama}, and FLAN-T5-Base \citep{FLANT5}); DP-Paraphrase \citep{dpparaphrase}, utilizing a GPT-2 model fine-tuned on SNLI; and DP-MLM \citep{dpmlm}, based on a RoBERTa-Base masked language model.

\begin{figure}[H]
  \centering
  % 第一行：Rouge1
  \begin{subfigure}{0.48\columnwidth}
      \centering
      \includegraphics[width=\linewidth]{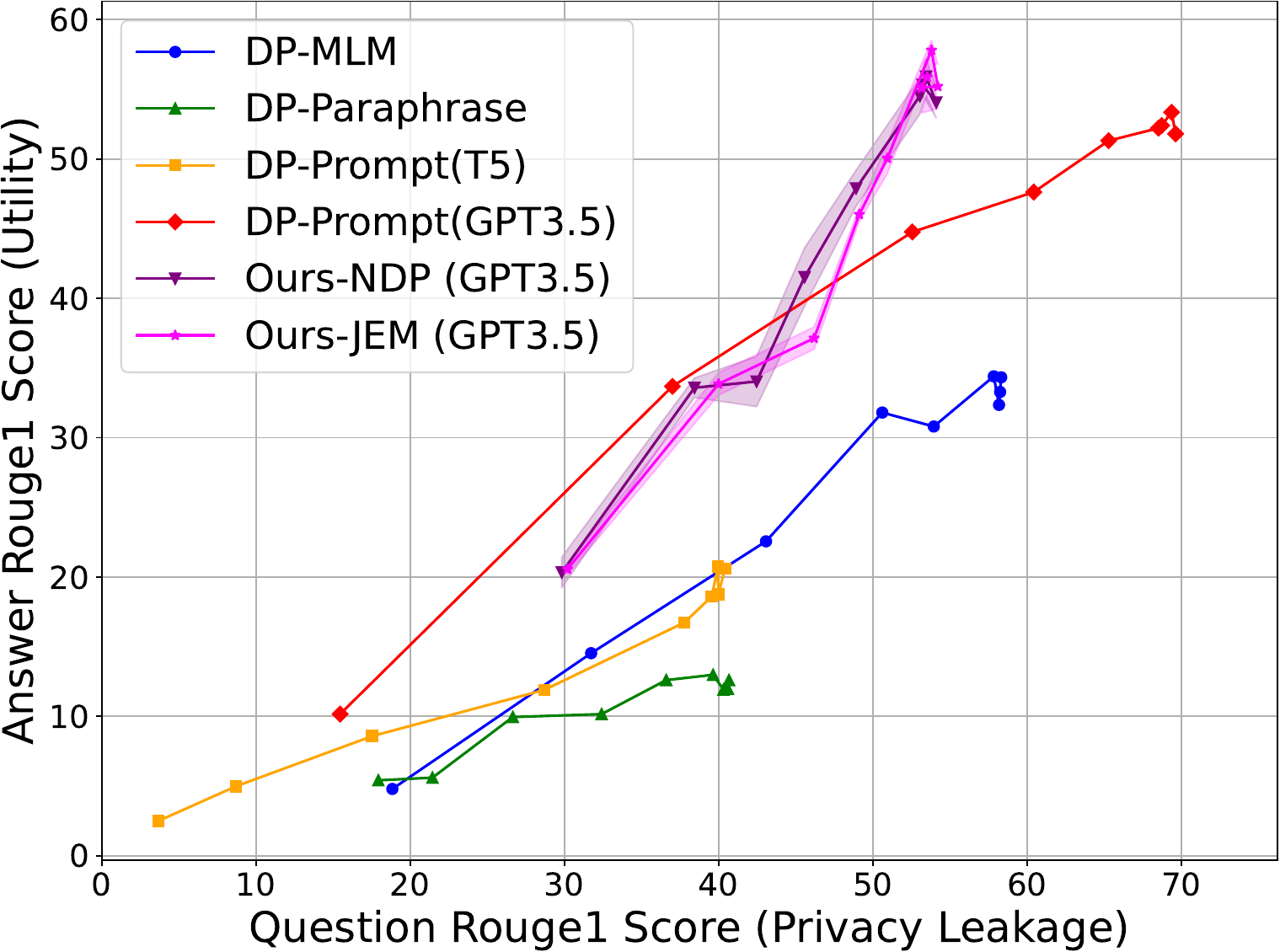}
      \caption{Rouge1 (GPT-3.5)}
  \end{subfigure}
  \hfill
  \begin{subfigure}{0.48\columnwidth}
      \centering
      \includegraphics[width=\linewidth]{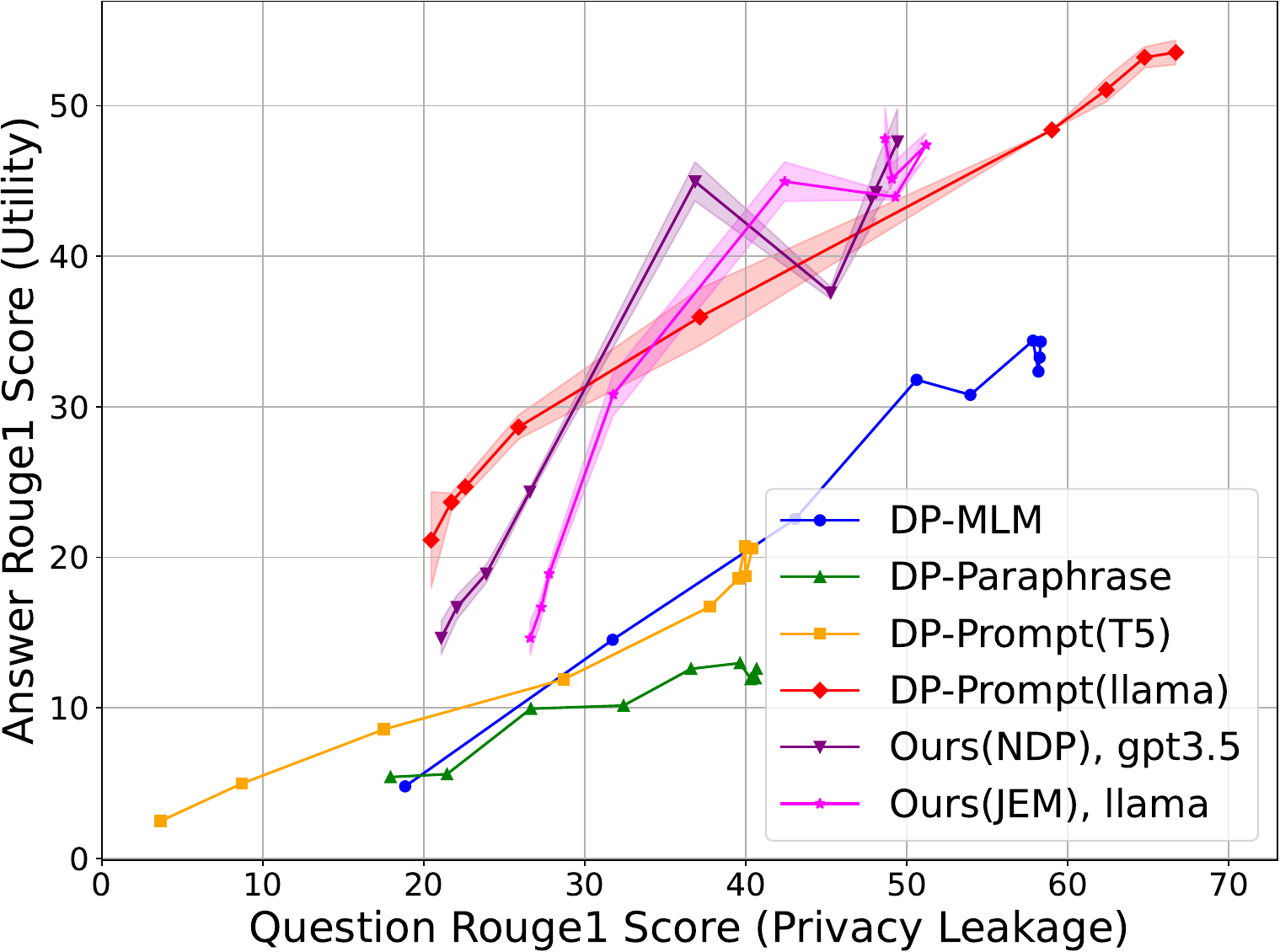}
      \caption{Rouge1 (Llama)}
  \end{subfigure}
  
  \vspace{1ex} % 调整行间距
  
  % 第二行：RougeL
  \begin{subfigure}{0.48\columnwidth}
      \centering
      \includegraphics[width=\linewidth]{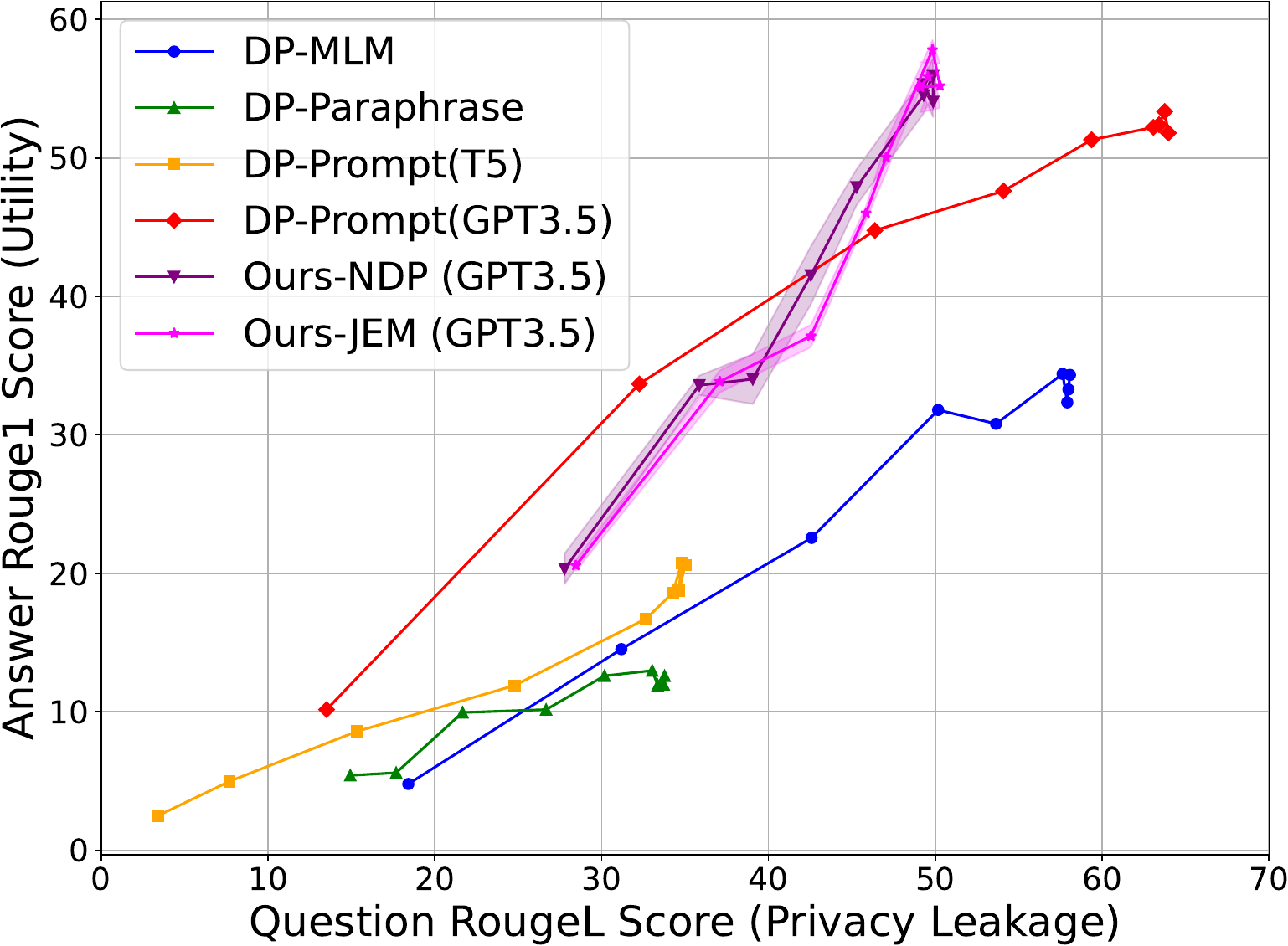}
      \caption{RougeL (GPT-3.5)}
  \end{subfigure}
  \hfill
  \begin{subfigure}{0.48\columnwidth}
      \centering
      \includegraphics[width=\linewidth]{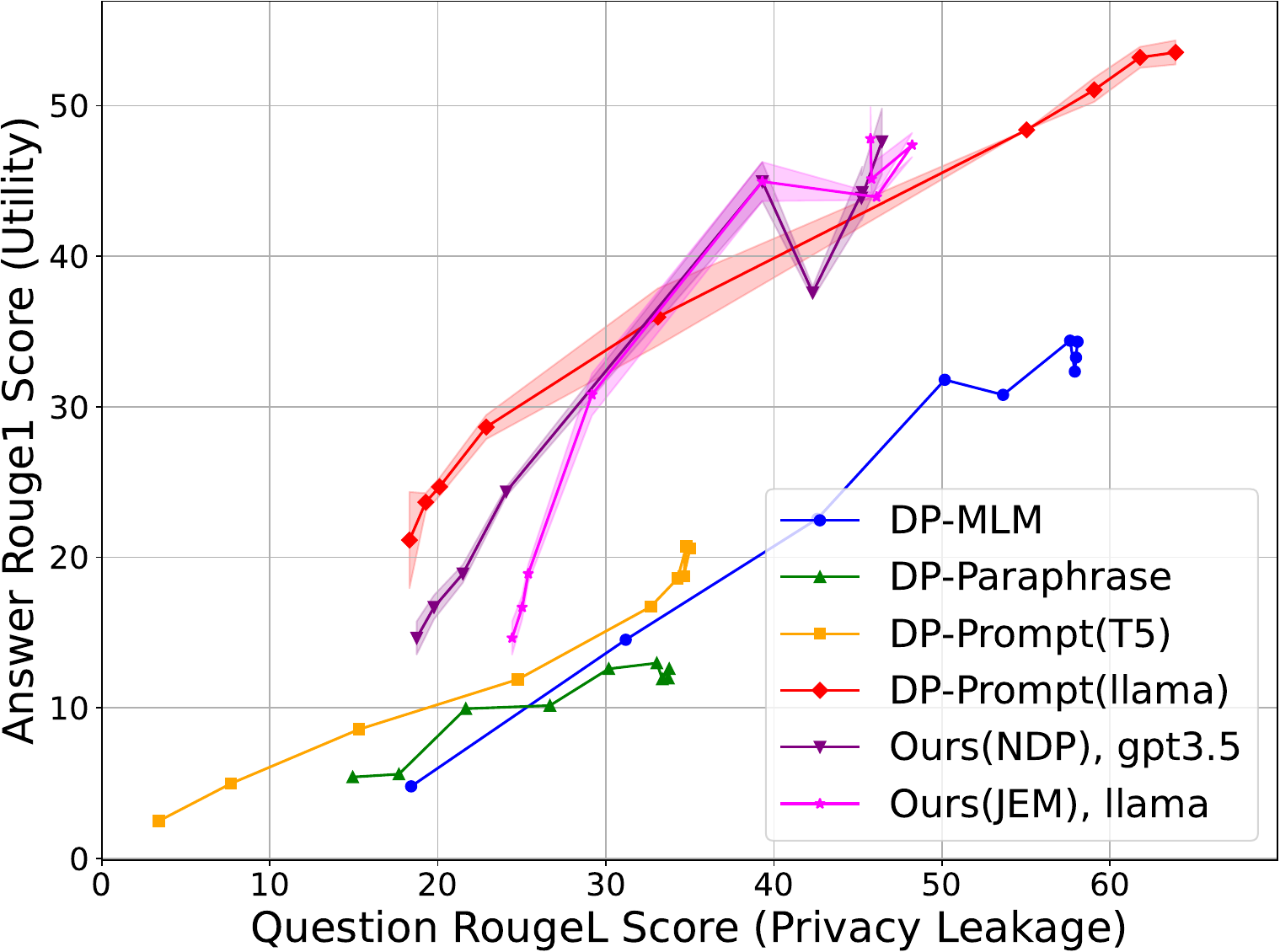}
      \caption{RougeL (Llama)}
  \end{subfigure}
  
  \vspace{1ex} % 调整行间距
  
  % 第三行：BLEU
  \begin{subfigure}{0.48\columnwidth}
      \centering
      \includegraphics[width=\linewidth]{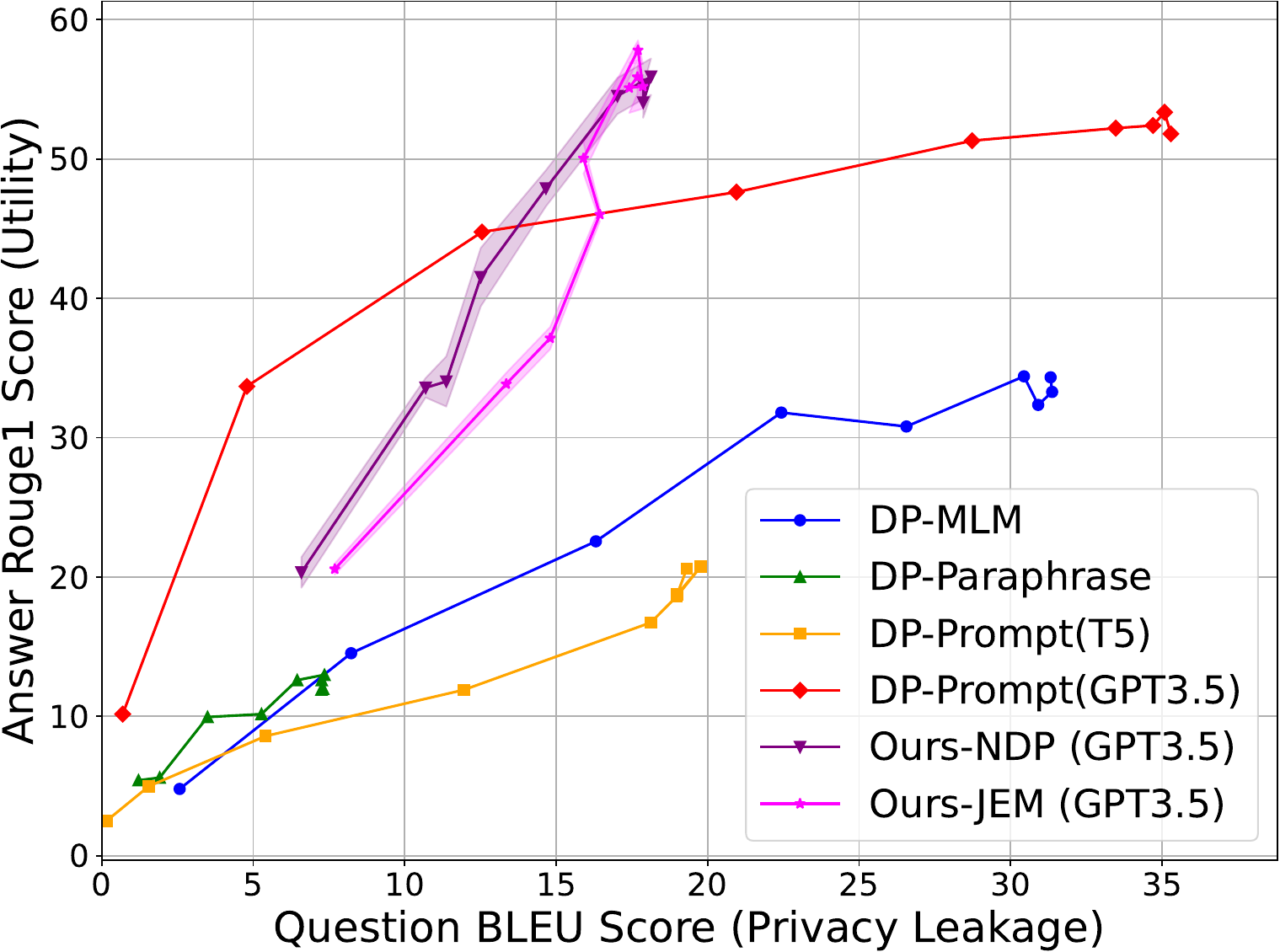}
      \caption{BLEU (GPT-3.5)}
  \end{subfigure}
  \hfill
  \begin{subfigure}{0.48\columnwidth}
      \centering
      \includegraphics[width=\linewidth]{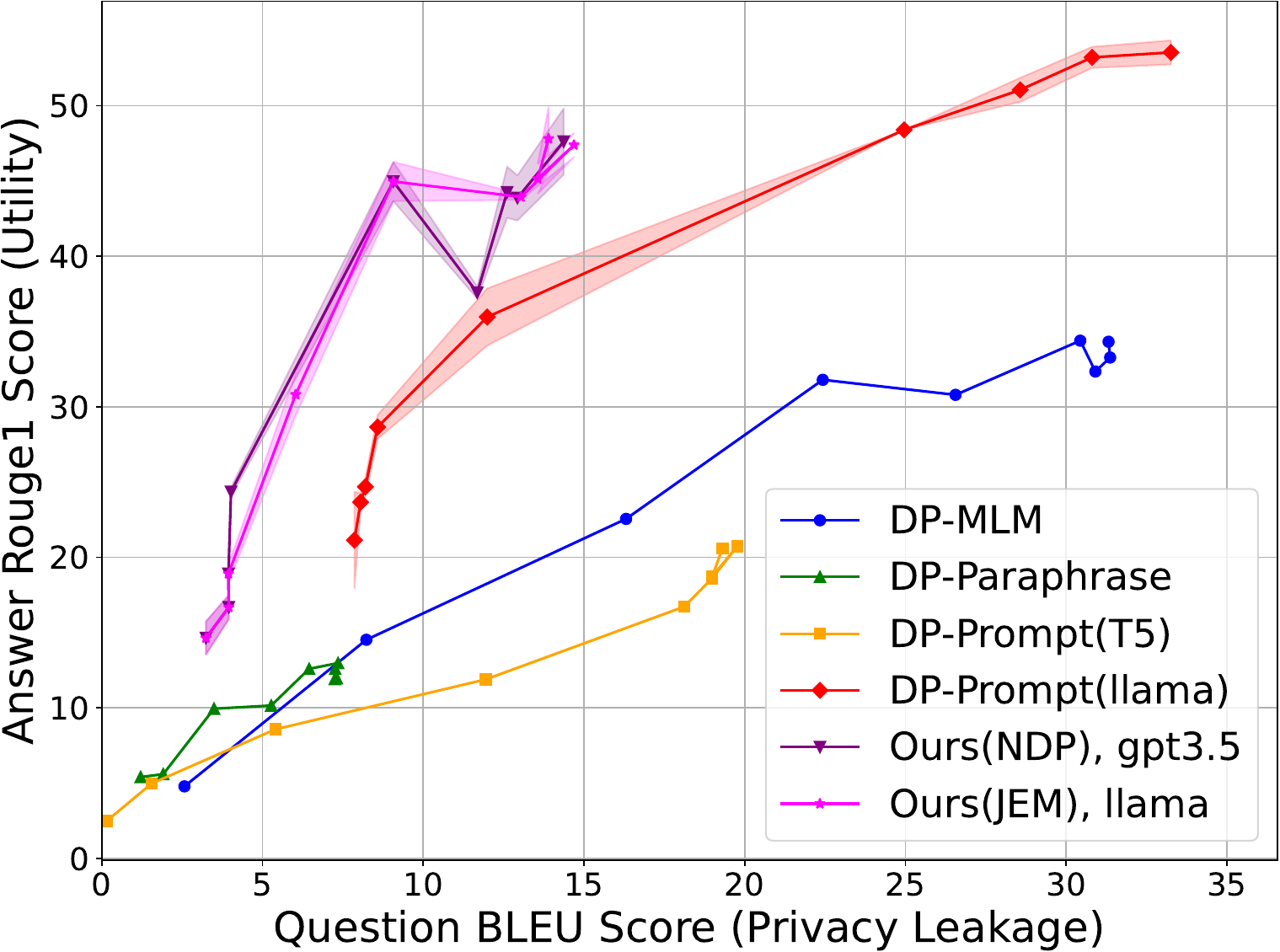}
      \caption{BLEU (Llama)}
  \end{subfigure}
  
  \caption{Privacy-utility trade-off for baselines and \modelname on open-answer PFL-DocVQA (VQA) dataset. The left column presents GPT-3.5 results, and the right column shows Llama-3.1-8B results. Refer to the x-axis label for specific measurement metrics.}
  \label{fig:res_vqa}
\end{figure}

\noindent\textbf{\modelname Settings.} We use GPT-3.5 (black-box) and Llama-3.1-8B (white-box) as underlying models for \modelname. Both the number of group text rewritings and private keywords are set to 10. For private keyword release in Stage-2, we implement a non-DP post-processing method (\textbf{\textit{\modelname-NDP}}) and a differentially private JointEM mechanism (\textbf{\textit{\modelname-JEM}}). In Stage-1, paraphrasing is controlled by temperature (black-box) and privacy budget $\epsilon$ (white-box), with nine values tested: $T \in \{0.1, 0.15, \dots, 1.5\}$. Corresponding $\epsilon$ values for the white-box model are calculated based on temperature and pre-clipped sensitivity (see Appendix Section~\ref{sec:eps}). Stage-3 prompt generation uses a temperature of 0 (or equivalently low) without a DP mechanism.

\begin{figure}[H]
  \centering
  % 第一行：Rouge1
  \begin{subfigure}{0.48\columnwidth}
      \centering
      \includegraphics[width=\linewidth]{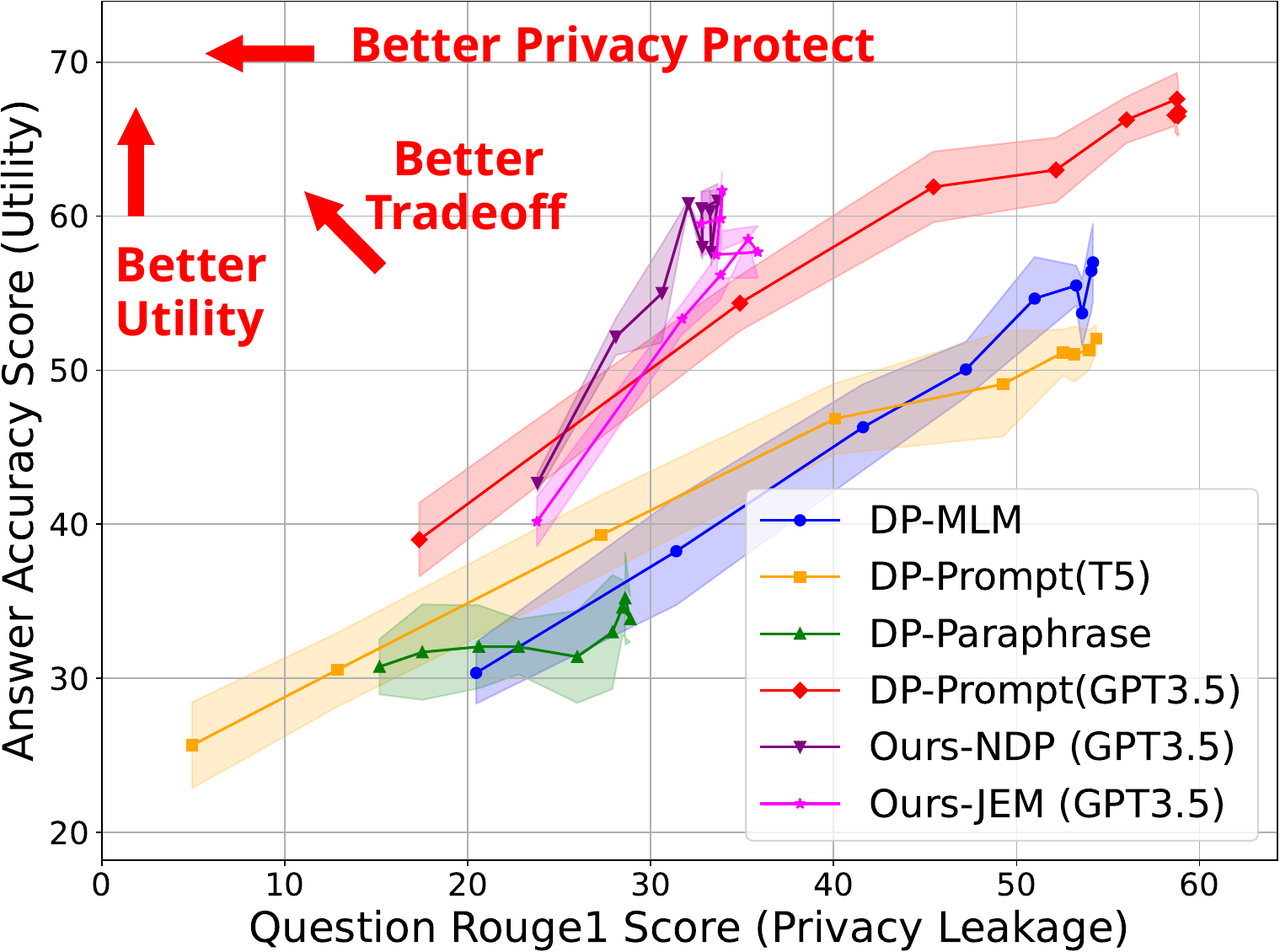}
      \caption{Rouge1 (GPT-3.5)}
  \end{subfigure}
  \hfill
  \begin{subfigure}{0.48\columnwidth}
      \centering
      \includegraphics[width=\linewidth]{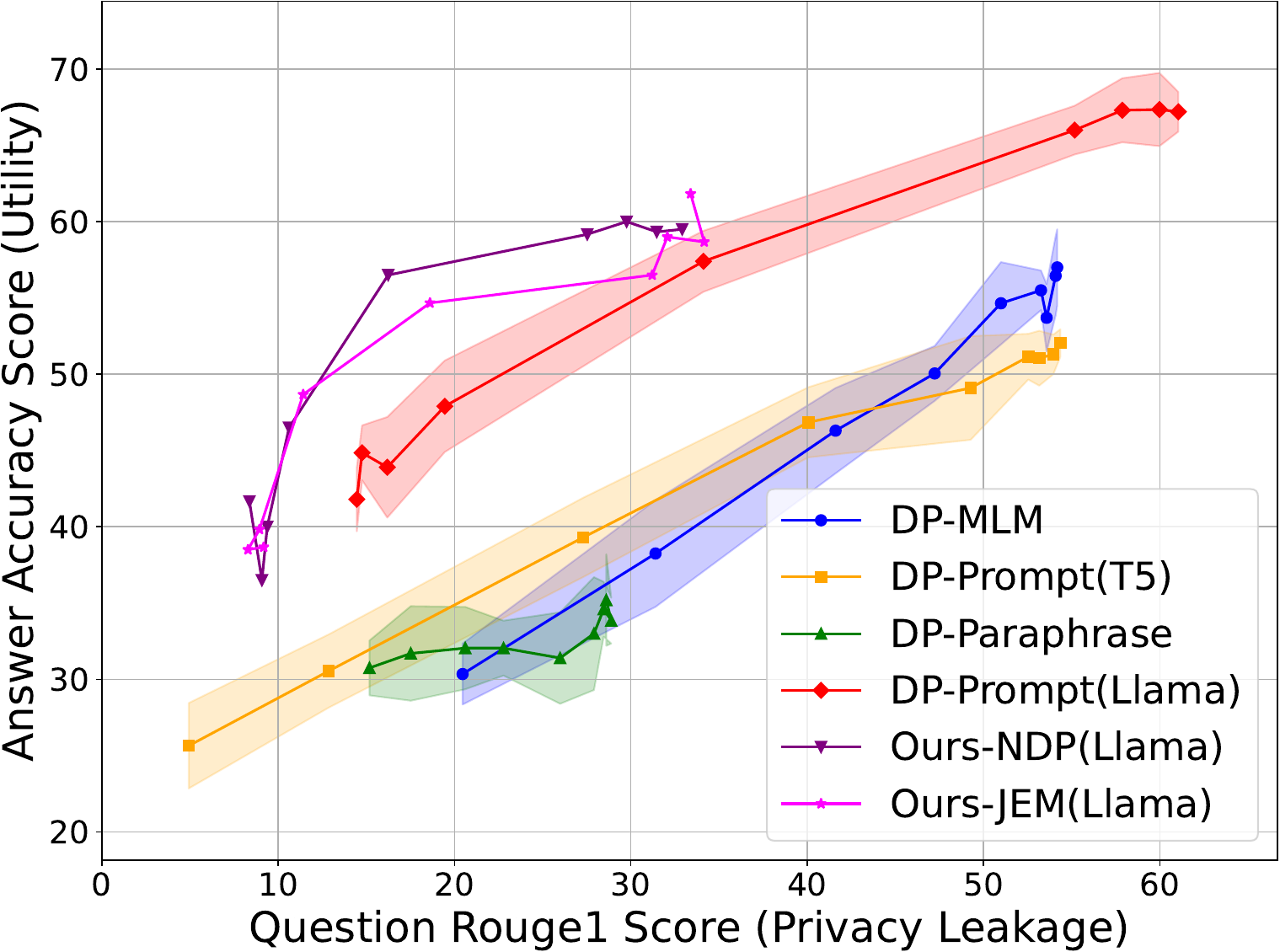}
      \caption{Rouge1 (Llama)}
  \end{subfigure}
  
  \vspace{1ex} % 行间距
  
  % 第二行：RougeL
  \begin{subfigure}{0.48\columnwidth}
      \centering
      \includegraphics[width=\linewidth]{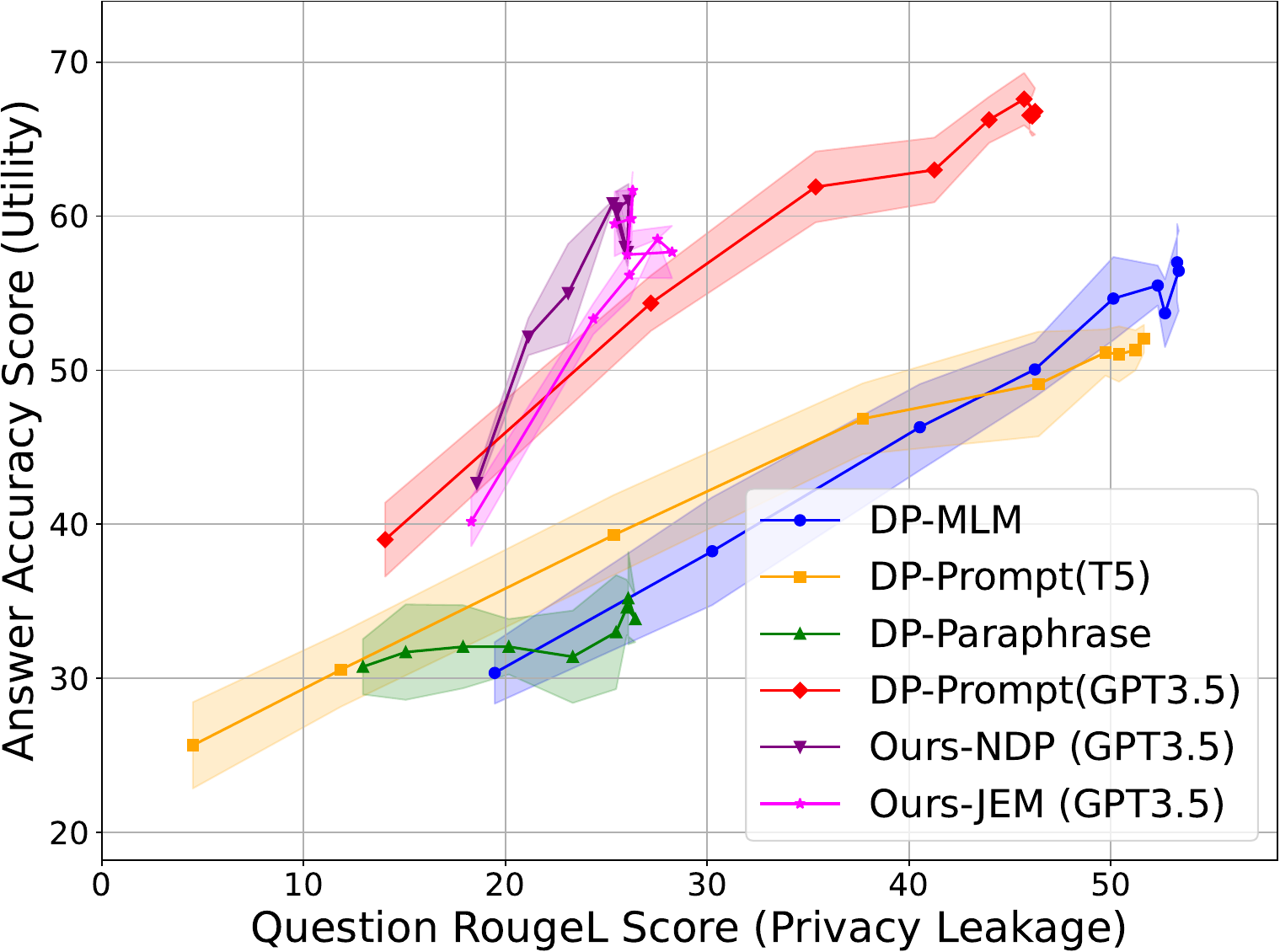}
      \caption{RougeL (GPT-3.5)}
  \end{subfigure}
  \hfill
  \begin{subfigure}{0.48\columnwidth}
      \centering
      \includegraphics[width=\linewidth]{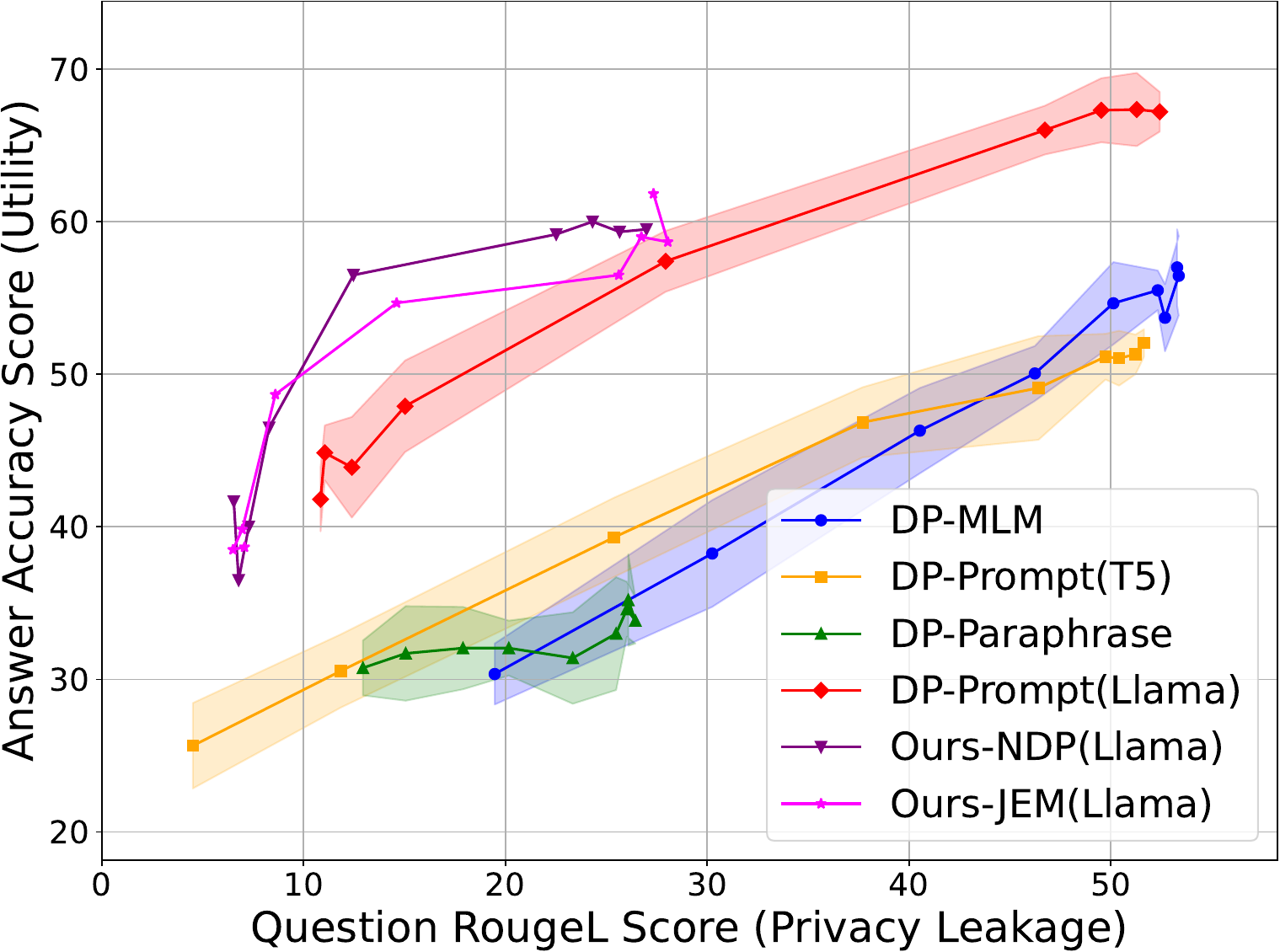}
      \caption{RougeL (Llama)}
  \end{subfigure}
  
  \vspace{1ex} % 行间距
  
  % 第三行：BLEU
  \begin{subfigure}{0.48\columnwidth}
      \centering
      \includegraphics[width=\linewidth]{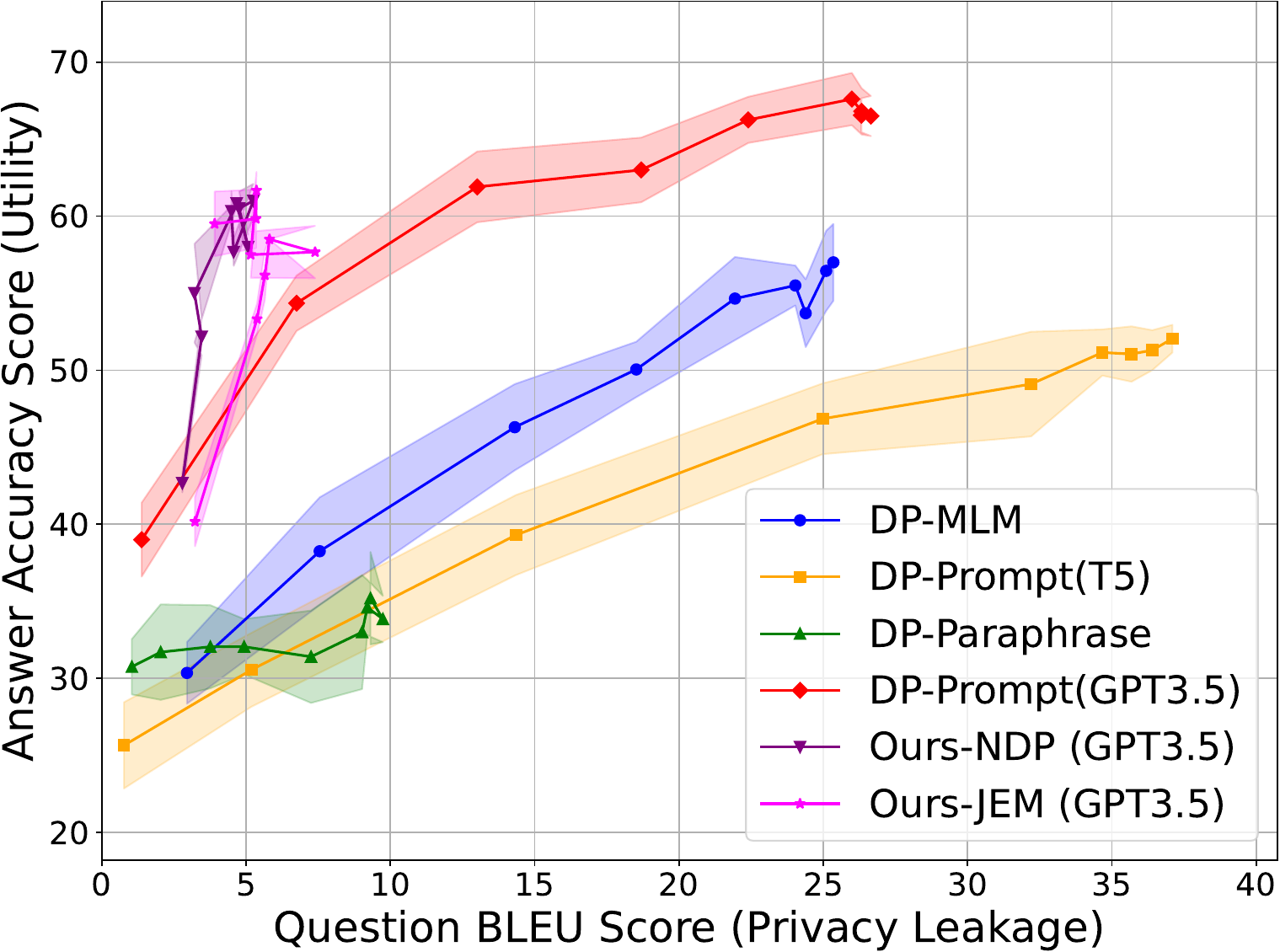}
      \caption{BLEU (GPT-3.5)}
  \end{subfigure}
  \hfill
  \begin{subfigure}{0.5\columnwidth}
      \centering
      \includegraphics[width=\linewidth]{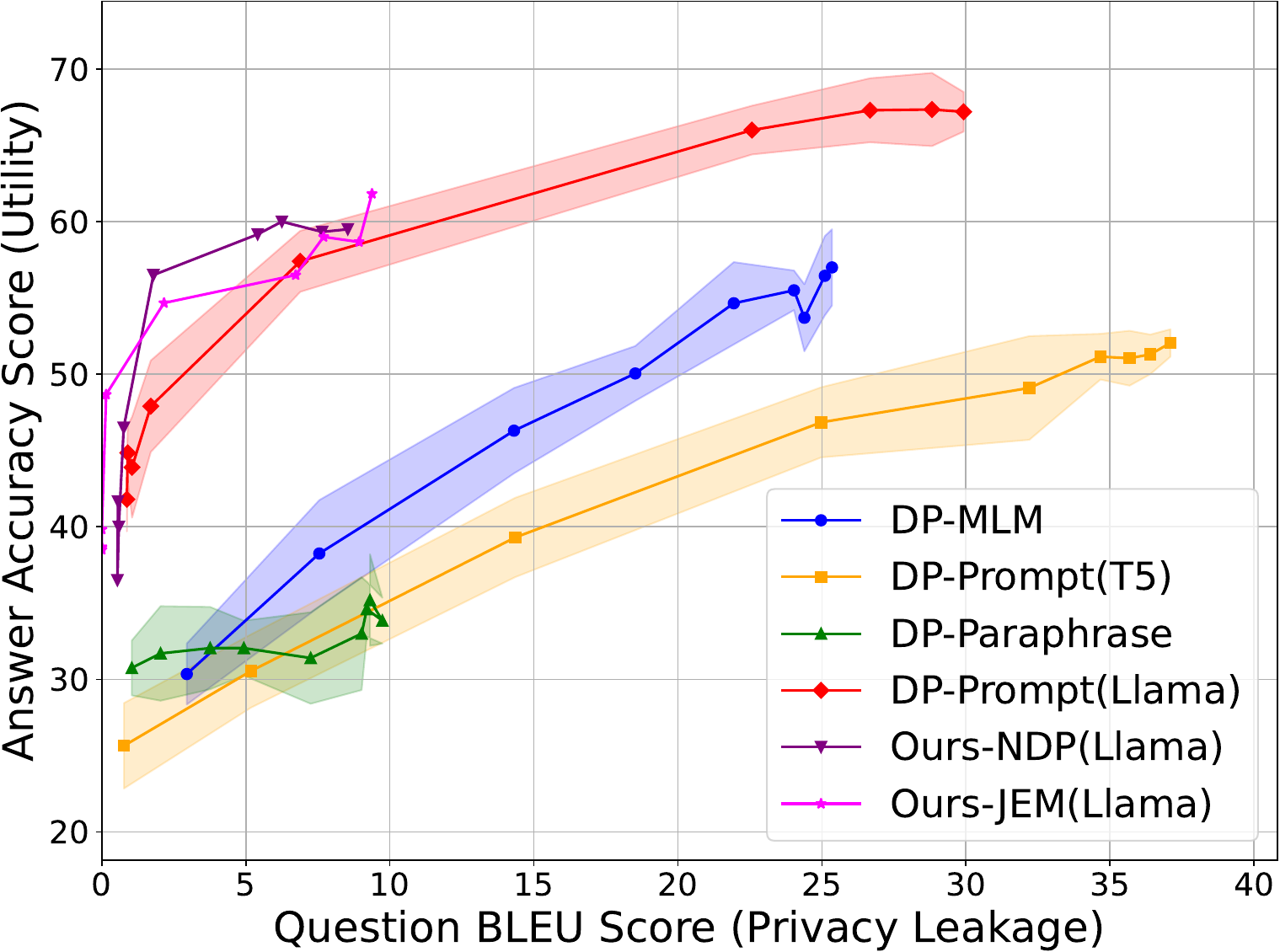}
      \caption{BLEU (Llama)}
  \end{subfigure}
  
  \caption{Privacy-utility trade-off for baselines and \modelname on close-answer Commonsense QA (CSQA) dataset. The left column presents GPT-3.5 results, and the right column shows Llama-3.1-8B results. Refer to the x-axis label for specific measurement metrics.}
  \label{fig:res_csqa}
\end{figure}

\noindent\textbf{Evaluation Repetitions.} All experiments were repeated five times, and the reported results are the mean values with standard deviations (displayed as shaded areas in Figures~\ref{fig:res_vqa} and~\ref{fig:res_csqa}).

\subsection{Results on Open-answer VQA Dataset}
Figure~\ref{fig:res_vqa} shows the results on the VQA dataset. \modelname (GPT-3.5) achieves a superior privacy-utility trade-off, consistently offering better privacy than DP-Prompt (GPT-3.5) at comparable utility, and sometimes higher utility at lower privacy. This validates our one-shot ICL utility design. With Llama, \modelname also maintains the best trade-off as temperature increases. \modelname-JEM, due to noisy keyword release, shows slightly higher privacy leakage compared to \modelname-NDP, but both provide strong privacy. Other baselines (DP-Prompt(T5), DP-MLM, DP-Paraphrase) achieve high privacy but at the cost of unacceptably low utility, demonstrating the limitations of prior evaluations relying on simplistic semantic analysis for utility in current LLM-based QA systems.

\subsection{Results on Close-answer CSQA Dataset}
Figure~\ref{fig:res_csqa} shows results on the CSQA dataset. \modelname converges faster and achieves a superior privacy-utility trade-off than baselines, generally outperforming them at equivalent privacy levels. While DP-Prompt (GPT-3.5 and Llama) shows higher utility in some cases, this comes at the cost of unacceptable privacy leakage. We identify a \textit{Rapid Equilibrium Deterioration Interval (REDI)}, where privacy degrades sharply with minor utility gains. DP-Prompt's REDI is wide and discrete (16-45\% for GPT-3.5, 20-55\% for Llama on question Rouge1), making parameter tuning difficult. \modelname mitigates this, converging around a question Rouge1 of 30\% and utility of 55-60\%, achieving a more stable and robust trade-off. The early convergence of \modelname on CSQA, a dataset with strong logical coherence, indirectly confirms the effectiveness of our privacy keyword suppression.

\subsection{Non-Uniform Rewriting Strategies}
Beyond uniform temperature or epsilon settings, we investigated the impact of \textit{non-uniform} rewriting strategies during group text rewriting. We conducted 10 rewriting tasks using DP-Prompt \citep{dpprompt} and our method (\modelname) on the GPT-3.5 model, with temperatures \(T\) ranging from 0.5 to 1.5 in increments of 0.1. Table~\ref{tab:combined} illustrates that our method achieves a favorable privacy-utility trade-off. For VQA, \modelname-NDP reduces privacy leakage from 56.91\% to 43.94\% while incurring a 2.69\% utility loss. For CSQA, privacy leakage decreases from 56.91\% to 43.94\%, with a corresponding 8.55\% utility loss. The total privacy budget in this non-uniform setting is  \(\sum_{i=1}^{10} \varepsilon_i\), where \(\varepsilon_i = \frac{2n\Delta u}{T_i}\),  \(T_i \in \{0.5, 0.6, \dots, 1.5\}\), and \(n\) is the number of tokens in the \(i\)-th generated text.

\begin{table}[!h]
\centering
\caption{Performance of rewriting under different temperatures on the VQA and CSQA datasets. 
Values are reported as mean (standard deviation) over five runs, for both Question (Privacy Leakage) and Answer (Utility).}
\label{tab:combined}
\resizebox{\linewidth}{!}{
\begin{tabular}{lccc c}
\toprule
\multicolumn{5}{c}{\textbf{Open-answer VQA Results}} \\
\midrule
\textbf{Methods} & \multicolumn{3}{c}{\textbf{Question (Privacy Leakage)}} & \textbf{Answer (Utility)} \\
\cmidrule(lr){2-4} \cmidrule(lr){5-5}
& \textbf{Rouge1} & \textbf{RougeL} & \textbf{BLEU} & \textbf{Rouge1} \\
\midrule
DP-Prompt & 56.91 (18.0) & 51.54 (16.8) & 23.24 (13.3) & 45.87 (0.0) \\
NDP       & 43.94 (1.3)  & 40.22 (1.1)  & 13.40 (0.6)  & 43.18 (1.3) \\
JEM       & 47.28 (0.8)  & 44.15 (0.5)  & 20.98 (0.5)  & 43.07 (0.7) \\
\midrule
\multicolumn{5}{c}{\textbf{Close-answer CSQA Results}} \\
\midrule
\textbf{Methods} & \multicolumn{3}{c}{\textbf{Question (Privacy Leakage)}} & \textbf{Answer (Utility)} \\
\cmidrule(lr){2-4} \cmidrule(lr){5-5}
& \textbf{Rouge1} & \textbf{RougeL} & \textbf{BLEU} & \textbf{Accuracy} \\
\midrule
DP-Prompt & 49.73 (13.5) & 38.66 (10.6) & 19.18 (9.4)  & 62.65 (8.3) \\
NDP       & 30.01 (0.3)  & 22.87 (0.2)  &  4.61 (0.1)  & 54.10 (1.0) \\
JEM       & 35.11 (1.1)  & 27.35 (1.2)  &  8.33 (0.8)  & 53.60 (1.7) \\
\bottomrule
\end{tabular}
}
\end{table}

\subsection{Generalizable Plug-in Framework}
\label{subsec:generalizable}
A significant contribution of this work is the development of a generalizable framework that functions as a plug-in, compatible with any existing paraphrasing method. The modular design, indicated by the blue arrow representing Stage-1 in Figure~\ref{fig:framework}, enables the replacement of our DP-based text rewriting component with alternative paraphrasing techniques. To demonstrate this generalizability, we integrated the strong baseline method, DP-Prompt \citep{dpprompt}, prior to our method, using the same base models (GPT-3.5 and Llama) in sequence. Thus, the results presented in Figures~\ref{fig:res_vqa} and \ref{fig:res_csqa} also validate the plug-in capabilities and efficiency of our framework.

\section{Ablation Study}
The ablation experiments are conducted on two additional configurations: one with references but without keywords, and another with keywords but without references. The configuration is GPT-3.5 and DocVQA. The results are shown in Tab.~\ref{tab:privacy_utility_ablation}.

\begin{table}[H]
\centering
\caption{Question (Privacy Leakage) and Answer (Utility) ablation study. Subscripts indicate the difference between \textit{without Reference/Keywords} and Ours-NDP. The privacy and utility metric is Rouge-1 (\%) between final prompt and original prompt, and Rouge-1 (\%) between query answer and ground-truth answer.}
\label{tab:privacy_utility_ablation}
\resizebox{0.488\textwidth}{!}{
\begin{tabular}{l|llll}
\toprule
\textbf{Temperature} & $T=0.1$ & $T=0.25$ & $T=0.5$ & $T=1.0$ \\
\midrule
\multicolumn{5}{c}{\textbf{Question (Privacy Leakage: Rouge-1)}} \\
\midrule
Ours-NDP          & 53.21 & 53.05 & 48.91 & 42.46 \\
w/o Reference     & 7.74\textsubscript{\textcolor{mred}{↓45.47}} & 8.12\textsubscript{\textcolor{mred}{↓44.93}} & 8.45\textsubscript{\textcolor{mred}{↓40.46}} & 9.44\textsubscript{\textcolor{mred}{↓33.02}} \\
w/o Keywords      & 50.11\textsubscript{\textcolor{mred}{↓3.10}} & 49.36\textsubscript{\textcolor{mred}{↓3.69}} & 48.05\textsubscript{\textcolor{mred}{↓0.86}} & 47.05\textsubscript{\textcolor{mgreen}{↑4.59}} \\
\midrule
\multicolumn{5}{c}{\textbf{Answer (Utility: Rouge-1)}} \\
\midrule
Ours-NDP          & 55.31 & 54.51 & 47.89 & 34.03 \\
w/o Reference     & 1.01\textsubscript{\textcolor{mred}{↓54.30}} & 1.04\textsubscript{\textcolor{mred}{↓53.47}} & 1.13\textsubscript{\textcolor{mred}{↓46.76}} & 1.57\textsubscript{\textcolor{mred}{↓32.46}} \\
w/o Keywords      & 21.61\textsubscript{\textcolor{mred}{↓33.70}} & 23.08\textsubscript{\textcolor{mred}{↓31.43}} & 22.76\textsubscript{\textcolor{mred}{↓25.13}} & 24.58\textsubscript{\textcolor{mred}{↓9.45}} \\
\bottomrule
\end{tabular}
}
\end{table}

\begin{table*}[!ht]   % 也可以用 [!b] 或 [!ht]
\centering
\caption{The results of the adversarial attack. The Rouge-1 (\%) scores measuring question similarity between the original and DP-GTR-processed prompts (\textbf{Ours}), as well as under Static and Adaptive attacks, across different temperatures for both NDP and JEM. Subscripts indicate change relative to ours.}
\label{tab:attack_main}
\resizebox{.8\textwidth}{!}{%
\begin{tabular}{l|llll}
\toprule
\textbf{Temperature} & $T=0.1$ & $T=0.25$ & $T=0.5$ & $T=1.0$ \\
\midrule
Ours-NDP           & 53.19 & 53.04 & 48.95 & 42.43 \\
Static Attack      & 52.27\textsubscript{\textcolor{mred}{↓0.92}} & 52.37\textsubscript{\textcolor{mred}{↓0.67}} & 48.31\textsubscript{\textcolor{mred}{↓0.64}} & 40.21\textsubscript{\textcolor{mred}{↓2.22}} \\
Adaptive Attack    & 49.87\textsubscript{\textcolor{mred}{↓3.32}} & 49.54\textsubscript{\textcolor{mred}{↓3.50}} & 45.01\textsubscript{\textcolor{mred}{↓3.94}} & 36.93\textsubscript{\textcolor{mred}{↓5.50}} \\
\midrule
Ours-JEM           & 53.50 & 53.77 & 50.92 & 46.17 \\
Static Attack      & 51.87\textsubscript{\textcolor{mred}{↓1.63}} & 52.67\textsubscript{\textcolor{mred}{↓1.10}} & 50.50\textsubscript{\textcolor{mred}{↓0.42}} & 43.68\textsubscript{\textcolor{mred}{↓2.49}} \\
Adaptive Attack    & 49.60\textsubscript{\textcolor{mred}{↓3.90}} & 50.48\textsubscript{\textcolor{mred}{↓3.29}} & 46.90\textsubscript{\textcolor{mred}{↓4.02}} & 39.77\textsubscript{\textcolor{mred}{↓6.40}} \\
\bottomrule
\end{tabular}%
}
\end{table*}

Experimental results indicate that using either isolated privacy keywords or the reference alone yields significantly lower performance than our approach. Removing the reference substantially impairs the model’s ability to generate valid questions, where utility's Rouge-1 score is no more than 1\%. Omitting private keywords disrupts the trade-off, resulting in slight privacy leakage that is disproportionate to the utility loss. Optimal performance is achieved only when both elements are combined. We attribute this to the fact that, although privacy keywords serve as suppression targets, LLMs can leverage these keywords to produce effective paraphrases, thereby enhancing utility.

\section{Adversarial Attack}
Based on the threat model in~\Cref{sec:prelim}, we design two adversarial experiments: \textbf{Static} and \textbf{Adaptive} attacks. In the static setting, the attacker is unaware of \modelname and directly attempts to recover the original question from cloud-received paraphrases. In the adaptive setting, the attacker fully understands \modelname, replicates its framework, and leverages privacy consensus keywords to infer the original question. In \Cref{tab:attack_main}, using GPT-3.5 and DocVQA as examples, we report the Rouge-1 (\%) score between the original and DP-GTR-processed prompts under different releasing strategies, along with the corresponding $\Delta$ for each attack. See detailed attack settings in Appendix~\ref{sec:attack_setting}.

The results of adversarial attacks demonstrate that \modelname offers strong robustness against both static and adaptive threats, with no observed increase in question Rouge-1 similarity, indicating no privacy leakage. These findings underscore effectiveness in safeguarding against adversarial risks.

\section{Conclusion}
This paper proposes \modelname, a novel three-stage local differential privacy (LDP) framework that leverages differentially private paraphrasing and the composition theorem through group text rewriting to enhance the privacy-utility trade-off. \modelname is the first approach, to our knowledge, to apply in-context learning for LDP prompt privatization and to connect global and local DP mechanisms via grouped paraphrased text.  Furthermore, our framework is generalizable and compatible with any existing paraphrasing technique. 
Evaluations on open- and closed-answer QA datasets (DocVQA and Commonsense QA), simulating real-world LLM application scenarios, demonstrate that \modelname achieves a significantly superior privacy-utility trade-off compared to existing state-of-the-art methods.  
With the rapidly increasing adoption of LLMs, \modelname provides a practical, robust and readily deployable solution for mitigating the risk of user prompt privacy leakage.

\section{Limitations}
The primary limitation of our work is that LLMs may sometimes fail to follow instructions, potentially leading to privacy leakage or an inability to learn from one-shot utility exemplars. In future work, an important direction is to shift control from outlier prompting to an internal LLM generation configuration. This approach will fundamentally address the issue of prompt failure in LLMs.

Another limitation arises from computing resource constraints, which led us to choose the Llama-3.1-8B open-source model. This model may not effectively learn from its prompt, resulting in relatively poor performance. We believe that with a more capable open-source model, the framework would perform better.

\section{Acknowledgment}
This work was supported in part by the CAHSI–Google Institutional Research Program (IRP) Grant and by the National Science Foundation (NSF) under grants CCF-2447834, HSI-2225229, CNS-2231519, and DUE-2225229.

\bibliography{custom}

\appendix

% \section{Appendix}
% \label{sec:appendix}

\section{DP-guaranteed Paraphrasing Proof}
\label{sec:proof}
\textbf{Proof.} Let $D$ and $D'$ represent two arbitrary documents, and $\mathbf{u}, \mathbf{u}' \in \mathbb{R}^{|V|}$ denote the associated logit vectors. Set all the logits lie within the interval $\,[b_{min},\, b_{max}]$ under sensitivity clipping. For a given token $v \in V$, let $i$ denote its position index and $u_i$ represent its corresponding logit from $\mathbf{u}$. Consequently, we obtain~\citep{dpprompt},

\resizebox{0.9\linewidth}{!}{$
\begin{aligned}
\frac{\text{Pr}[M(D) = v]}{\text{Pr}[M(D') = v]} &= \frac{\frac{\exp(\frac{u_i}{T})}{\sum_{j=1}^{|V|} \exp(\frac{u_j}{T})}}{\frac{\exp(\frac{u_i'}{T})}{\sum_{j=1}^{|V|} \exp(\frac{u_j'}{T})}} \\
&= \frac{\exp(\frac{u_i}{T})}{\exp(\frac{u_i'}{T})} \frac{\sum_{j=1}^{|V|} \exp(\frac{u_j'}{T})}{\sum_{j=1}^{|V|} \exp(\frac{u_j}{T})} \\
&= \exp\left(\frac{u_i - u_i'}{T}\right) \frac{\sum_{j=1}^{|V|} \exp(\frac{u_j'}{T})}{\sum_{j=1}^{|V|} \exp(\frac{u_j}{T})} \\
&\le \exp\left(\frac{b_{max} - b_{min}}{T}\right) \exp\left(\frac{b_{max} - b_{min}}{T}\right) \\
&\le \exp\left(\frac{2(b_{max} - b_{min})}{T}\right).
\end{aligned}
$}

\begin{algorithm}[h]
\caption{DP-Prompt Algorithm~\citep{dpprompt}}
\label{alg:dp_prompt}
\begin{algorithmic}[1]
\REQUIRE Language model $\text{LM}$, Private document $D$, Private Budget $\epsilon$, Prompt template $T$, Logit bounds $\mathbf{b} \in \mathbb{R}^{|\mathcal{V}|}$ with $b_{min} \le u_{v} \le b_{max}$,  Number of generated tokens $n$
\ENSURE Sanitized text $P$

\STATE \textbf{Generate Prompt:} 
  Construct an initial context $\tilde{C}$ from $\{D, T\}$ and tokenize it
\STATE $\text{LM} \gets \texttt{clipLogits}(u, \mathbf{b})$
\STATE $\text{Temp} \gets \bigl(\tfrac{2\bigl(b_{max}-b_{min}\bigr)}{\epsilon}\bigr)$
\STATE $\text{LM} \gets \texttt{setTemperature}(\text{Temp})$
\FOR{$i = 1$ to $n$}
  \STATE $u \gets \text{LM}(\tilde{C})$ 
  \STATE $v \gets \texttt{ExponentialMechanism}(u)$
  \STATE $P \gets P \cup \{v\}$,\quad $\tilde{C} \gets \tilde{C} \cup \{v\}$
\ENDFOR

\STATE \textbf{Output:} Detokenize($P$)

\end{algorithmic}
\end{algorithm}

\section{Epsilon and Sensitivity Setting}
\label{sec:eps}
The sensitivity bound is the other critical theoretical parameter. Following prior work~\cite{igamberdiev2023dpbart}, we adopt a pre-clipping strategy. Specifically, we randomly sample 1,000 examples from the CSQA training dataset and perform the DP-Prompt paraphrasing task while recording all logits. We then compute the mean ($\mu$) and standard deviation ($\sigma$), and define the sensitivity bound as $(\mu, \mu+4\sigma)$ to better preserve high-value logits~\citep{dpmlm}. We also detail the exact logits clipping ($[b_{min}, b_{max}]$) range used in~\Cref{tab:b_range}.

\begin{table}[htbp]
\label{tab:b_range}
\centering
\caption{Logits clipping bounds ($b_{min}$ and $b_{max}$) for different DP-based models.}
\resizebox{0.488\textwidth}{!}{
\begin{tabular}{lcc}
\toprule
\textbf{Model} & $b_{min}$ & $b_{max}$ \\
\midrule
DP-MLM (RoBERTa)         & -3.2093  & 16.3048  \\
DP-Paraphrase (GPT-2)    & -96.8525 & -8.7477  \\
DP-Prompt (Flan-T5)      & -19.2271 & 7.4832   \\
Ours (Llama-3.1-8B)      & 22.9443  & 32.6274  \\
\bottomrule
\end{tabular}
}
\end{table}

The corresponding \(\epsilon\) is computed using the alignment target temperature with the formula \((\frac{2(b_{\max}-b_{\min})}{T})\). See the Table~\ref{tab:epsilon} for detailed values.

\begin{table}[htbp]
\centering
\caption{Epsilon values for different methods introduced in~\Cref{sec:exp_setup} across temperatures (\textbf{T}).}
\resizebox{0.488\textwidth}{!}{
\begin{tabular}{c c c c c}
\toprule
\multirow{2}{*}{\textbf{T}} & \textbf{DP-MLM} & \textbf{DP-Paraphrase} & \textbf{DP-Prompt} & \textbf{Ours} \\
\cmidrule(lr){2-5}
 & \textbf{RoBERTa} & \textbf{GPT-2} & \textbf{FLAN-T5} & \textbf{Llama} \\
\midrule
0.10  & 390.0  & 1760.0 & 534.2  & 194.0 \\
0.15  & 260.0  & 1173.3 & 356.1  & 129.3 \\
0.20  & 195.0  & 880.0  & 267.1  & 97.0  \\
0.25  & 156.0  & 704.0  & 213.7  & 77.6  \\
0.50  & 78.0   & 352.0  & 106.8  & 38.8  \\
0.75  & 52.0   & 234.7  & 71.2   & 25.9  \\
1.00  & 39.0   & 176.0  & 53.4   & 19.4  \\
1.25  & 31.2   & 140.8  & 42.7   & 15.5  \\
1.50  & 26.0   & 117.3  & 35.6   & 12.9  \\
\bottomrule
\end{tabular}
}
\label{tab:epsilon}
\end{table}

Additionally, in the setting of Ours-JEM, $\varepsilon_2$ for JointEM is set to \textbf{2}.

\section{Computational Costs}
\modelname is an inference-side, single-prompt privacy protection approach that demonstrates lower overall computational resource consumption compared to methods requiring model training~\cite{dpopt}. The primary source of computational overhead stems from the grouping step involved in paraphrasing. Below, we provide a detailed computational analysis.

\begin{itemize}
    \item \textbf{Computation Efficiency:} \modelname is designed to support parallel generation of paraphrased texts, which can substantially reduce inference time from $O(n+1)$ to $O(2)$ when there are sufficient computational resources.
    \item \textbf{Memory Efficiency:} When parallel processing is not feasible, GTR can operate in serial mode. In this case, only the sentence with the lowest perplexity and the corresponding privacy keyword votes are retained in memory, thereby minimizing memory consumption from $O(n)$ to $O(1)$.
    \item \textbf{User-End Efficiency:} For user-end environments with low throughput and limited computational resources, \modelname does not require model training and offers particular advantages in both memory and computation efficiency.
\end{itemize}

While we acknowledge that \modelname introduces additional computational overhead relative to baseline inference, we argue that this trade-off is justified. The enhanced privacy protection offered by \modelname is particularly important for practical deployments, given the increasing privacy risks associated with large language models.

\section{Adversarial Attack}
\label{sec:attack_setting}
We consider a threat model in which the attacker has full capability at the cloud end at Preliminaries~\ref{sec:prelim}. Accordingly, we detail both static and adaptive attack strategies. In the static attack, the adversary directly prompts the LLM to reconstruct the original question using the template shown in Template~\ref{blk:static}. For the adaptive attack, the adversary replicates the entire workflow, including candidate paraphrase generation, private consensus keyword extraction, and reference selection, and then combines this information to recover the original question. The adaptive prompt recovery template is provided in Template~\ref{blk:adaptive}.

\refstepcounter{mybox}\label{blk:static}
\begin{tcolorbox}[colback=gray!4!white, colframe=black!40!white, title=Static Prompt Template, width=0.5\textwidth]
Given a paraphrased question: <$\hat{P}$> 

Convert to the original question: 
\end{tcolorbox}

\refstepcounter{mybox}\label{blk:adaptive}
\begin{tcolorbox}[colback=gray!4!white, colframe=black!40!white, title=Adaptive Prompt Template, width=0.5\textwidth]
Refer the paraphrased question to generate the original question:  
<\(\hat{P}_{low}\)>  

Use following tokens:  
<\(\hat{w}_1, \hat{w}_2, ..., \hat{w}_k\)>

Generated original question:
\end{tcolorbox}

\section{Explainable Case}
\label{sec:case}

To illustrate our approach, we present an explainable case below, which is forked from our open-source repository and generated by \modelname-NDP~(GPT-3.5). The \textbf{Document} is the original question, followed by \textbf{10 DP-guaranteed paraphrases}. From these, we extract the \textbf{Private Keywords} and identify the lowest-perplexity \textbf{Reference}, ultimately producing a differentially private, protected document. 

\begin{tcolorbox}[
  colback=gray!4!white,
  colframe=black!40!white,
  title=\modelname\ Example
]
\textbf{Document:}

\texttt{In which year, john f. kennedy was assassinated?}
\\

\textbf{10 DP-guaranteed Paraphrases:}
\begin{itemize}[leftmargin=0.7em, itemsep=0pt, topsep=0pt]
    \item In what year was John F. Kennedy killed?
    \item What year was John F. Kennedy assassinated?
    \item During which year did John F. Kennedy get assassinated?
    \item In what year was John F. Kennedy killed?
    \item What year did John F. Kennedy get assassinated?
    \item In what year was John F. Kennedy killed?
    \item What year did John F. Kennedy get assassinated?
    \item In what year was John F. Kennedy killed?
    \item What year did John F. Kennedy get shot?
    \item In what year did John F. Kennedy get killed?\\
\end{itemize}

\textbf{Private Keywords:} \\
{\ttfamily What, get, Kennedy, killed, In, John, assassinated, F., year, During}
\\

\textbf{Reference:} \\
{\ttfamily What year was John F. Kennedy assassinated?}
\\

\textbf{Final Generated Document:} \\
{\ttfamily When did the tragic event occur involving the 35th President of the United States?}
\end{tcolorbox}

\end{document}